\documentclass{article}

\usepackage[preprint]{neurips_2026}

\usepackage[utf8]{inputenc} 
\usepackage[T1]{fontenc}    
\usepackage{hyperref}       
\usepackage{url}            
\usepackage{booktabs}       
\usepackage{amsfonts}       
\usepackage{nicefrac}       
\usepackage{microtype}      
\usepackage{xcolor}         
\usepackage{graphicx}
\usepackage{enumitem}       
\usepackage{multirow}       
\usepackage{wrapfig}        
\usepackage{colortbl}       
\usepackage{array}          

\usepackage{subcaption}
\usepackage{wrapfig}
\usepackage{multicol}
\usepackage{adjustbox}

\usepackage{amsmath}
\usepackage{amssymb}
\usepackage{mathtools}
\usepackage{amsthm}

\let\para\paragraph

\title{Valence–Arousal Subspace in LLMs: Circular Emotion Geometry and Multi-Behavioral Control}

\author{%
  Lihao Sun \\
  University of Chicago \\
  \texttt{slhleosun@uchicago.edu}
  \And
  Lewen Yan \\
  Shanghai AI Lab
  \And
  Xiaoya Lu \\
  Shanghai AI Lab
  \And
  Andrew Lee \\
  Harvard University
  \And
  Jie Zhang \\
  Shanghai AI Lab
  \And
  Jing Shao \\
  Shanghai AI Lab \\
  \texttt{shaojing@pjlab.org.cn}
}

\begin{document}

\maketitle

\begin{abstract}
We show that emotion vectors in LLMs are organized by a two-dimensional valence--arousal (VA) subspace exhibiting circular geometry. Through principal component decomposition and ridge regression, we recover meaningful VA axes underlying emotion steering vectors whose projections correlate with human affect ratings across 44,728 words. Steering along these axes produces monotonic control over the affective properties of generated text, and further affords bidirectional control over multiple downstream behaviors (refusal and sycophancy) from a single subspace. These effects replicate across Llama~3.1-8B, Qwen3-8B, and Qwen3-14B. We propose \emph{lexical mediation} to explain why these effects and prior emotionally framed controls work: refusal and compliance tokens occupy distinct VA regions, and VA steering directly modulates their emission probabilities.
\end{abstract}


\section{Introduction}

A growing body of work shows that emotionally framed prompting or activation steering influences large language model (LLM) behavior \citep{li2023largelanguagemodelsunderstand, konen2024stylevectorssteeringgenerative, reichman2025emotionsartthouunderstanding, dong2025rationalanswersemotionalresonance}. Yet why such methods work---and when they fail---remains unclear.

One obstacle is conceptual.
Work that studies such phenomena in LLMs typically treats discrete emotion categories---\textit{anger, joy, fear}---as fundamental units of analysis \citep{zhang2024refashioning,hollinsworth2024language,plazadelarco2024emotionanalysisnlptrends, konen2024stylevectorssteeringgenerative, dong2025rationalanswersemotionalresonance}.
To find a common basis for comparison, a growing trend borrows from human psychology the two-dimensional framework of valence (pleasure--displeasure) and arousal (activation--deactivation) \citep{ishikawa2025aiemotionsexploringemotional, felixpena2025emotionalframing}.
In practice, these dimensions are often equated with categorical labels or used merely as evaluation metrics.
Though yielding insights, this discreteness-driven perspective forgoes the explanatory power that the original distinction affords.

In the psychological tradition from which these constructs originate, valence and arousal (VA) do not define emotions themselves but rather \textit{core affect}---a continuous experiential substrate from which discrete emotions emerge \citep{russell1980circumplex, russell2003coreaffect}. 
Emotions arise when core affect is attributed to a cause and labeled with a culturally available category: a pleasant, high-arousal state, attributed to an unexpected gift, becomes surprise or excitement. 
We invoke this distinction not as a claim about LLM phenomenology, but as a modeling choice: continuous dimensions and discrete labels play different explanatory roles and should not be conflated. 

In this work, we treat categorical labels and VA axes as separate structures and demonstrate the insights this decoupling affords. At the level of model representations, this conceptual shift also motivates a methodological one. Many existing steering-vector methods construct directions via contrastive differences (e.g., mean activation differences over paired examples) \citep{panickssery2024steeringllama2contrastive}. While effective, these directions can be brittle---often requiring careful tuning across prompts, layers, or behaviors, and may not directly transfer across tasks \citep{tan2025analyzinggeneralizationreliabilitysteering,braun2025understandingunreliabilitysteeringvectors,oozeer2025activationspaceinterventionstransferred}.
We show that decomposing emotion steering vectors into a low-dimensional VA subspace yields shared structure that generalizes across multiple downstream behaviors, suggesting a more reusable control basis than single-task contrastive directions.

\begin{figure*}[t]
    \centering
    \includegraphics[width=0.93\textwidth]{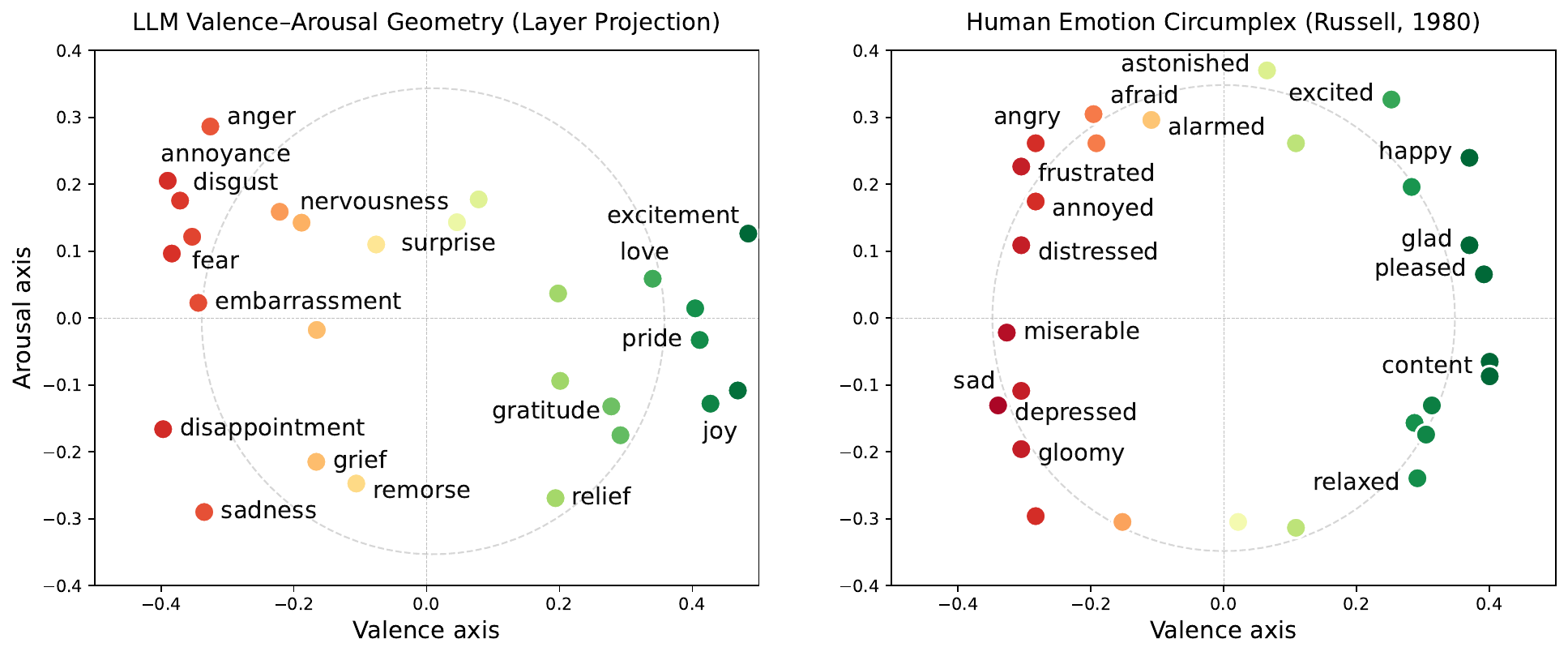}
    \caption{\textbf{Emotion steering vectors projected onto the VA subspace} at layer~31, colored by valence. Gray circle: algebraic least-squares fit. The circular arrangement is analogous to the circumplex model of affect in human psychology \citep{russell1980circumplex}.}
    \label{fig:circumplex}
    \vspace{-10pt}
\end{figure*}

Concretely, we derive emotion steering vectors from 211,225 emotion-labeled text samples~\citep{demszky2020goemotionsdatasetfinegrainedemotions} and learn VA subspaces via ridge regression over principal components. The resulting emotion projections exhibit circular geometry analogous to the circumplex model from human emotion perception \citep{russell1980circumplex}, as shown in Figure~\ref{fig:circumplex}. In addition, the axes correlate with human VA ratings across 44,728 words~\citep{mohammad2025nrcvadlexiconv2}. Steering along these axes produces monotonic control over the affective properties of model outputs, and further affords bidirectional control over refusal and sycophancy---with effects replicating across Llama~3.1-8B, Qwen3-8B, and Qwen3-14B. We propose \emph{lexical mediation} as one explanation: refusal and compliance behaviors have load-bearing signature tokens (e.g., ``can't,'' ``sorry'' for refusal; ``sure,'' ``Here'' for compliance) that occupy distinct VA regions, so VA steering shifts their emission probabilities, modulating downstream behavior. We show that prior emotion-based prompting methods induce corresponding VA shifts, suggesting a shared underlying mechanism.

Taken together, our work advances understanding of affective phenomena in LLMs by motivating a conceptual separation between discrete emotion labels and continuous valence--arousal dimensions. We identify a 2D subspace aligned with human-interpretable VA concepts and show that emotion vectors are systematically organized as a circumplex within this geometry. From an interpretability standpoint, we present a principal-component-based decomposition method for identifying interpretable subspaces underlying steering vectors, exposing shared structure that generalizes and supports multi-behavioral control in contrast to typical task-specific contrastive steering. Finally, we propose lexical mediation as one explanation and useful perspective for understanding why emotion-based controls work in LLMs.

\section{Related Work}
\textbf{Emotion-framed analysis and control in LLMs.}
Appending emotionally framed phrases to prompts influences LLM behavior across domains~\citep{li2023largelanguagemodelsunderstand,good_bad_why}. Prior work has investigated the internal structure underlying these effects, including hierarchical organization of emotion representations~\citep{zhao2025emergencehierarchicalemotionorganization}, emotion-specific neurons and attention heads~\citep{wang2025llmsfeelemotioncircuits}, and affective biases~\citep{NegativePrompt,zhou2024alignment}. However, these approaches center on emotion labels as discrete categories; we take a step further to find a meaningful underlying subspace. Concurrent work by Anthropic~\citep{sofroniew2026emotionconcepts} independently identifies emotion representations that form a similar circumplex geometry in Claude Sonnet 4.5. Our work decomposes these into a continuous 2D subspace and provides a unifying mechanistic account linking emotionally framed behaviors across representation geometry, unembedding structure, and neuron-level analyses.

\textbf{Geometry of representations in LLMs.}
A growing body of work characterizes structures in representations. Many studies focus on single directions that predict or control a behavior~\citep{turner2024steeringlanguagemodelsactivation,panickssery2024steeringllama2contrastive,sun2025alignedblindalignmentincreases,lee2025geometryselfverificationtaskspecificreasoning}, including sentiment~\citep{tigges2023linearrepresentationssentimentlarge}. Two-dimensional geometries are less commonly characterized~\citep{nanda2023progressmeasuresgrokkingmechanistic, gurnee2024languagemodelsrepresentspace}. We add to this smaller body by demonstrating a 2D VA geometry for emotion representations with finer insights otherwise overlooked by a sentiment-only account.

\textbf{Activation steering.}
Activation steering enables training-free behavioral control by manipulating internal representations~\citep{zou2025representationengineeringtopdownapproach}, but typically relies on task-specific contrastive vectors for each target behavior~\citep{turner2024steeringlanguagemodelsactivation,qianetal2024towards,lee2025shared}. Emotions have been shown to correspond to linear directions in activation space~\citep{tigges2023linearrepresentationssentimentlarge,hollinsworth2024language}. We present an approach that uncovers meaningful subspaces underlying steering vectors, recovering interpretable, reusable axes for multi-behavioral control in contrast to the typical single-behavior scope of contrastive steering vectors.

\section{Identifying Valence and Arousal Subspaces}
\label{sec:identification}

We present a method to identify meaningful subspaces from steering vectors, which proceeds in three stages: (1) extracting emotion directions via mean-difference contrasts, (2) eliciting model self-reported VA coordinates for each emotion label, and (3) learning VA axes as linear combinations of principal components via ridge regression. We report results on Llama 3.1-8B-Instruct \citep{grattafiori2024llama3herdmodels}, Qwen3-8B, and Qwen3-14B \citep{qwen3} to assess cross-architecture generality.

\para{Emotion steering vectors.}
We first derive emotion steering vectors using contrastive means \citep{tigges2023linearrepresentationssentimentlarge,panickssery2024steeringllama2contrastive, dong2025rationalanswersemotionalresonance}. We use GoEmotions \citep{demszky2020goemotionsdatasetfinegrainedemotions}, which comprises 211,225 text samples annotated with 27 emotion labels plus a neutral class. We retain only examples annotated with exactly one emotion. For each emotion category $e$, we extract the last-token hidden state of each sample at every layer. The steering vector at layer $\ell$ is
\begin{equation}
    \mathbf{v}_e^{(\ell)} = \frac{1}{|D_e|} \sum_{x \in D_e} \mathbf{h}^{(\ell)}(x) - \frac{1}{|D_{\text{neutral}}|} \sum_{x \in D_{\text{neutral}}} \mathbf{h}^{(\ell)}(x)
\end{equation}
where $\mathbf{h}^{(\ell)}(x) \in \mathbb{R}^H$ denotes the last-token hidden state at layer $\ell$ for input $x$; $D_e$ and $D_{\text{neutral}}$ are the sets of single-label examples for emotion $e$ and the neutral category. 

Next, we obtain VA scores by prompting the model to self-report the valence and arousal for each emotion label (e.g., "anger," "joy," "sadness"), averaging across three prompt templates for robustness (see Appendix~\ref{app:self-report-prompt}). We operationalize both dimensions as continuous values in the range $[-1, +1]$: valence spans from extremely unpleasant ($-1$) to extremely pleasant ($+1$), and arousal spans from very calm ($-1$) to very activated ($+1$), with $0$ denoting neutrality on each axis.

\para{Decomposing steering vectors into subspaces.}
For each layer $\ell$, we first mean-center the emotion steering vectors $\mathbf{V}^{(\ell)} \in \mathbb{R}^{K \times H}$ and apply principal component analysis to obtain a low-rank basis. Let $\mathbf{U}_k \in \mathbb{R}^{H \times k}$ denote the top $k$ principal component directions, and let $\mathbf{Z} \in \mathbb{R}^{K \times k}$ denote the projection of the centered emotion vectors onto these components (i.e., the PC scores). We then fit ridge regression models to recover the valence and arousal scores:
\begin{equation}
    \hat{\boldsymbol{\beta}}_V = \arg\min_{\boldsymbol{\beta}} \|\mathbf{Z}\boldsymbol{\beta} - \tilde{\mathbf{y}}_V\|^2 + \lambda \|\boldsymbol{\beta}\|^2
\end{equation}
where $\tilde{\mathbf{y}}_V$ denotes the mean-centered valence ratings, and analogously for arousal. The learned coefficients $\hat{\boldsymbol{\beta}}_V \in \mathbb{R}^k$ define the valence axis as a linear combination of principal components. The corresponding direction in the original activation space is given by $\mathbf{w}_V = \mathbf{U}_k \hat{\boldsymbol{\beta}}_V$, normalized to unit length. With Gram--Schmidt orthogonalization, we ensure the V and A axes are orthonormal. 

\begin{figure}[t]
    \centering
    \includegraphics[width=0.85\columnwidth]{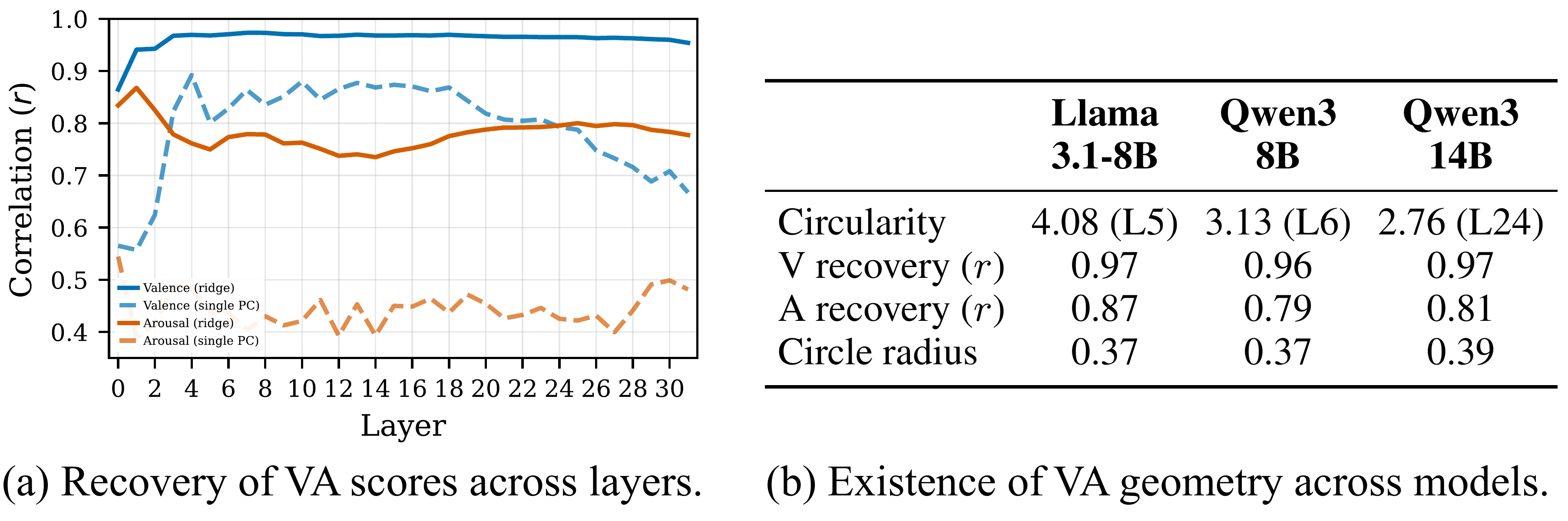}
    \caption{\textbf{VA subspace recovery and generalizability.}
    \textbf{(a)} Learned subspace projections recover target valence and arousal scores across layers.
    \textbf{(b)} Comparable circularity, recovery correlations, and circle radius across architectures suggest that VA subspaces exist across models.}
    \label{fig:va_recovery_generalizability}
    \vspace{-10pt}
\end{figure}

Figure~\ref{fig:va_recovery_generalizability}(a) reports recovery performance across layers. Using PC1 alone, valence is well captured ($r = 0.89$ at layer 4), with correlations exceeding $0.80$ across most middle layers---indicating valence aligns with the principal axis of variation among emotion steering vectors. Arousal, by contrast, is poorly captured by any single principal component ($r = 0.54$). Ridge regression over multiple components improves both: valence reaches $r = 0.97$ and arousal reaches $r = 0.87$. This asymmetry implies valence constitutes the dominant dimension of emotion steering vectors, while arousal is distributed across secondary components.

\para{Circular geometry of emotion representations.}
Projecting emotion steering vectors onto the learned VA subspace reveals a circular arrangement analogous to Russell's circumplex model found in human psychology~\citep{russell1980circumplex}. Qualitatively, positive emotions (\textit{joy, gratitude}) oppose negative ones (\textit{sadness, grief}) along valence, while high-arousal states (\textit{excitement, anger}) separate from low-arousal states (\textit{relief, sadness}) along the orthogonal axis (Figure~\ref{fig:circumplex}). To quantify this structure, we fit circles to the projected coordinates using algebraic least squares, minimizing squared radial residuals. 
We report circularity (the ratio of mean to standard deviation of distances from the fitted center), where higher values indicate tighter circular arrangement. Figure~\ref{fig:circumplex} shows the fit at layer~31, with similar circular geometry observed across layers.

We note a methodological subtlety: the VA axes are fitted on the model's self-reported scores for 27 emotion \emph{labels}, while the steering vectors projected onto these axes are mean-difference activations computed over GoEmotions \emph{text passages}. The regression, by construction, defines directions that align with how the model rates 27 words, and the circumplex structure of steering vectors projected on the label-derived axes is not strictly guaranteed by the method.

\para{Cross-model generalization.}
To assess whether the VA subspace is specific to Llama or reflects a more general property, we apply the same extraction and fitting pipeline to Qwen3-8B and Qwen3-14B. Figure~\ref{fig:va_recovery_generalizability}(b) reports the key metrics. The VA subspace consistently emerges across architectures: circle radii (0.37, 0.37, 0.39) and recovery correlations are all comparable, confirming that the geometric structure exists across models. 

Three further observations are worth distinguishing at this stage. 
First, replacing the model's self-reported VA scores with crowdsourced human ratings from the NRC-VAD lexicon~\citep{mohammad2025nrcvadlexiconv2} as regression targets yields comparable or improved recovery in all three models (arousal: Llama $0.87 \to 0.95$; Qwen3-8B $0.79 \to 0.83$; Qwen3-14B $0.81 \to 0.87$), while valence remains at $r = 0.97$. The axes recovered under human and self-report supervision are closely aligned ($|\cos| > 0.90$ for valence and $|\cos| > 0.65$ for arousal at all layers across all three models). This suggests that the models have learned to report VA ratings coherent with human annotations. 

Second, the alignment between label-derived axes and text-derived steering vectors indicates a degree of internal coherence. The model's behavioral reports about emotions (how it rates ``anger'' on VA) are consistent with the geometric structure of its own representations when processing emotion-laden text, suggesting a form of representational--behavioral coherence.

Third, while the identification process surfaces circular geometry, it alone does not establish that the identified subspaces truly correspond to human-interpretable concepts of valence and arousal or causally influence model behavior. We conduct intervention experiments in the following sections to test whether the VA subspaces are indeed tightly related to affective properties in LLMs. There, we focus on the self-report-supervised VA subspaces to test whether model-native affect structure---recovered without external human supervision---has genuine functional significance.

\section{Validation of VA Subspaces}
\label{sec:validation}
We now explore whether the recovered VA subspaces correspond to human-interpretable notions of valence and arousal, with two experiments: (1) whether word representations, when projected onto the subspaces, yield VA scores consistent with human ratings; and (2) whether steering the model along the VA axes leads to corresponding changes in the affective properties of generated text.

\subsection{Correlation with Human-Crowdsourced Lexicon Annotations}
We use the NRC-VAD Lexicon~\citep{mohammad2025nrcvadlexiconv2}, which contains human-crowdsourced VA ratings for 44,728 English words. For each word, we project the model's representation onto the VA subspace and compute the projection components along the V and A axes, then correlate these with the human scores. Concretely, we extract the representation at the last token position and project onto the VA axes using the same centering as in Section~\ref{sec:identification}: $v_{\text{proj}} = (\mathbf{h} - \boldsymbol{\mu}) \cdot \mathbf{w}_V$ and $a_{\text{proj}} = (\mathbf{h} - \boldsymbol{\mu}) \cdot \mathbf{w}_A$, where $\boldsymbol{\mu}$ is the mean activation computed during subspace fitting.

Valence projections correlate strongly with human ratings, reaching $r = 0.71$ ($\rho = 0.69$) at layer~6. Arousal projections show weaker correlations, peaking at $r = 0.23$ ($\rho = 0.22$) at layer~7. 
This asymmetry, where valence is more predictable than arousal from de-contextualized words, aligns with prior findings in affective computing~\citep{snefjella2016delivery,delatorre2019contextaffectivenorms,debruyne2021annotatingvad,mendes2023quantifyingvalencearousaltext,choi2026latentstructureaffectiverepresentations}: arousal depends on situational context that isolated words cannot provide, whereas valence is more stably encoded in lexical semantics. Both correlations are significant ($p < 10^{-16}$), supporting that the VA axes capture human-interpretable affective dimensions.

\begin{figure}[t!]
    \centering
    \includegraphics[width=0.94\textwidth]{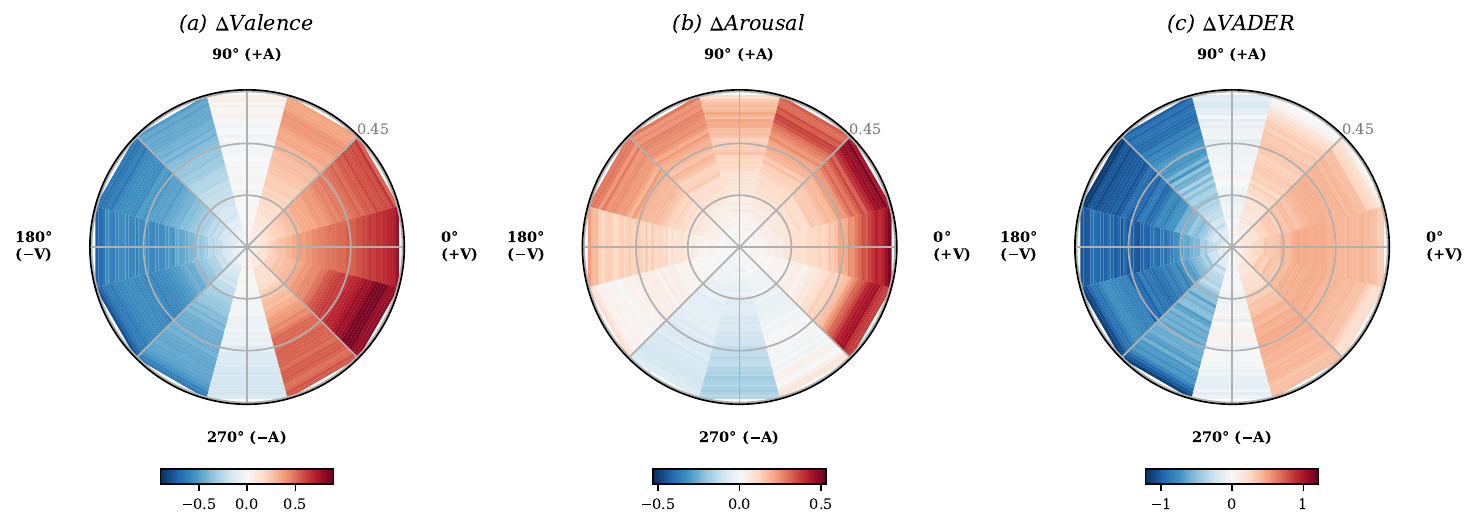}
    \caption{\textbf{Effects of VA steering on affective properties of open-ended generation.} Each radial heatmap shows the change of one affective property relative to the unsteered baseline across steering directions on the VA subspace (angle) and steering strengths (radius, $\alpha \in [0.01, 0.45]$). (a)~Valence change. (b)~Arousal change. (c)~Sentiment change.}
    \label{fig:steering_radial}
    \vspace{-10pt}
\end{figure}

\subsection{Controlling Affective Properties of Responses with VA Steering}
\label{sec:steering_validation}

Next, to establish the link between VA axes and affective properties, we steer models along directions spanning multiple angles in the VA subspace using open-ended prompts, and measure changes in the affective properties of model responses.

\para{Method.} We steer along 12 angular directions in VA space ($0^\circ, 30^\circ, \ldots, 330^\circ$), where $0^\circ$ aligns with positive valence and $90^\circ$ with positive arousal, at 45 steering strengths ($\alpha \in [0.01, 0.45]$). Steering is applied at every layer by adding the steering vector to the hidden state at every token position during greedy decoding. We evaluate on 130 open-ended prompts adapted and expanded from~\citet{konen2024stylevectorssteeringgenerative}, comprising emotionally neutral story continuations and factual questions (Appendix~\ref{app:validation_prompts}). The neutral prompt design isolates steering effects from prompt-induced emotion.
Generated responses are scored by two external systems trained for evaluating affective properties of text. VADER~\citep{hutto2014vader} is a lexicon-based sentiment analyzer that outputs a score in $[-1, +1]$, where $-1$ indicates maximally negative sentiment and $+1$ maximally positive. 
VAD-BERT~\citep{buechel2022emobankstudyingimpactannotation,robrokools2022vadbert} is a regressor trained on the EmoBank corpus that outputs valence and arousal scores on a 1--5 scale, where 1 indicates low and 5 high. Unsteered baseline generations occupy a near-neutral region of affect space (mean valence $= 3.05$, mean arousal $= 3.29$ on the 1--5 VAD-BERT scale), providing room for bidirectional modulation. Figure~\ref{fig:steering_radial} shows the change in each metric relative to the unsteered baseline as a function of steering direction (angle) and strength (radius).

\para{Valence steering produces clear, symmetric effects on response affect.} As shown in Figure~\ref{fig:steering_radial}(a), the $0^\circ$ direction yields $\Delta V = +0.75$ at maximum strength, while $180^\circ$ produces $\Delta V = -0.73$, spanning 1.5 points on the VAD-BERT scale. VADER sentiment closely tracks this pattern (Figure~\ref{fig:steering_radial}(c)), confirming that the learned V direction captures positive--negative affect consistent with human-interpretable concepts.

\para{Arousal steering modulates intensity with minimal valence leakage.} As shown in Figure~\ref{fig:steering_radial}(b), the $90^\circ$ direction increases arousal ($\Delta A = +0.50$), while $270^\circ$ decreases it ($\Delta A = -0.16$). Notably, arousal steering produces negligible changes in valence or VADER sentiment (Figure~\ref{fig:steering_radial}(a) and (c)), indicating that the V and A axes control separable aspects of affect: the model's emotional intensity can be modulated independently of its positive--negative tone.

We also observe collateral arousal increases from valence steering: both $+V$ ($0^\circ$: $\Delta A = +0.49$) and $-V$ ($180^\circ$: $\Delta A = +0.22$) elevate arousal. This reflects a well-documented property of affective language: highly valenced content---whether extremely pleasant or unpleasant---tends to be rated as more arousing than valence-neutral content~\citep{warriner2013norms}, so steering toward stronger valence naturally induces more arousing text.

Taken together, these results validate that the VA subspace captures functionally meaningful affective structure: valence steering bidirectionally controls positive--negative affect, arousal steering independently modulates emotional intensity, and the two axes exhibit the separability expected of orthogonal affective dimensions. These findings suggest that the VA axes identified in Section~\ref{sec:identification} are not merely geometric abstractions but are linked to controllable affective properties of model output.

\begin{figure*}[t!]
    \centering
    \includegraphics[width=0.9\textwidth]{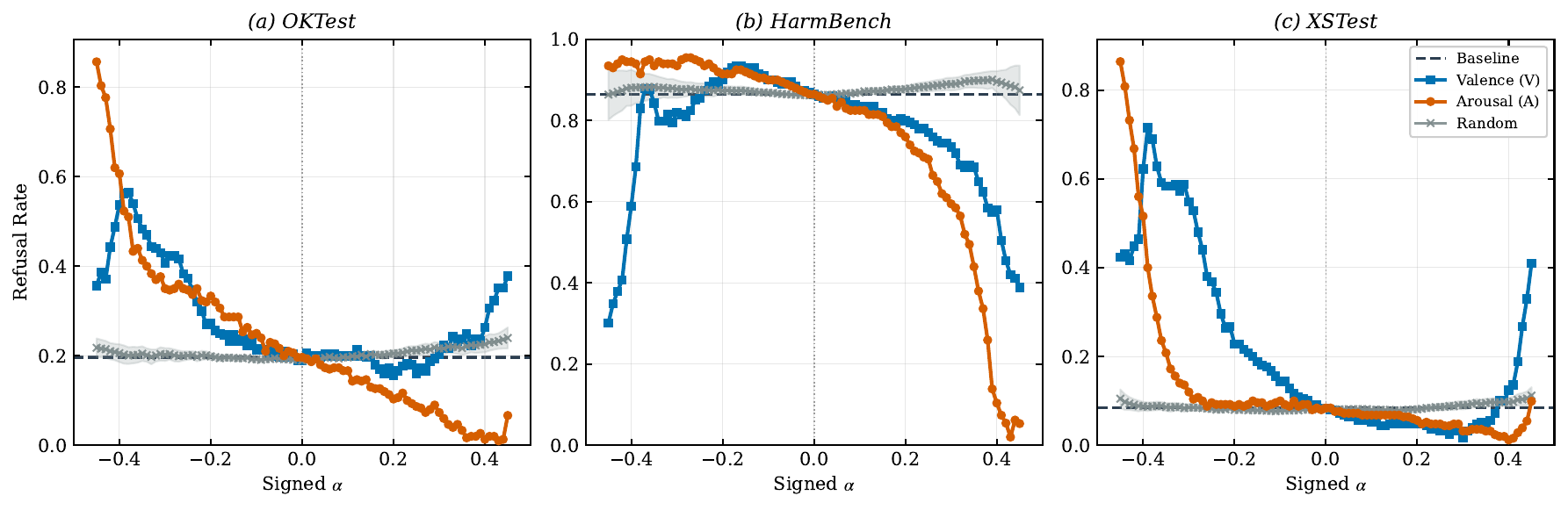}
    \caption{\textbf{VA steering controls refusal behavior.} Refusal rate as a function of signed steering strength $\alpha$ across three benchmarks. Arousal steering (orange) provides clean, near-monotonic bidirectional control. Random directions (gray) produce no systematic effect.}
    \label{fig:refusal_interventions}
    \vspace{-8pt}
\end{figure*}

\section{Controlling Refusal and Sycophancy with a Single VA Subspace}
\label{sec:interventions}

The preceding sections established that VA axes capture affective structure at both the lexical and generative levels. Next, we explore whether this affective substrate interacts with higher-level model behaviors. Here, we conduct VA steering experiments on refusal and sycophancy.

We use the same steering methodology as in Section~\ref{sec:steering_validation}, applied along the V and A axes separately. As controls, we steer along three categories of random directions (in-plane, orthogonal-to-plane, and fully random orthonormal pairs; 3 seeds each). All nine controls produce similarly flat effects.

\subsection{VA for Refusal Control}
\label{sec:refusal}
We test the effect of VA steering on refusal across three benchmarks: OKTest~\citep{shi2024navigatingoverkilllargelanguage}, HarmBench~\citep{mazeika2024harmbenchstandardizedevaluationframework}, and XSTest~\citep{röttger2024xstesttestsuiteidentifying}. Refusal is evaluated using Ai2 WildGuard~\citep{wildguard2024}, trained for refusal and harmfulness detection in LLM responses.

\para{Arousal provides clean, bidirectional control over refusal.}
As shown in Figure~\ref{fig:refusal_interventions}, steering along the arousal axis yields monotonic control over refusal across all three benchmarks. Decreasing arousal consistently increases refusal (OKTest $20\% \to 86\%$ at $\alpha = -0.45$; XSTest $8\% \to 86\%$), while increasing arousal suppresses it (HarmBench $87\% \to 5\%$ at $\alpha = +0.45$). Random control directions remain within $2$--$3\%$ of baseline across all values of $\alpha$ tested. 

Valence steering also modulates refusal, but with less consistent directionality. On XSTest, the effect is directionally clean ($8\% \to 55\%$ at $\alpha = -0.3$; $8\% \to 2\%$ at $+0.3$) but less consistent across other benchmarks at the same steering strengths. These results suggest that arousal and valence play different roles in refusal mediation, a dissociation that would not be visible from a single emotion or sentiment direction, and that is surfaced by the VA decomposition.

We report results for $\alpha \in [-0.45, 0.45]$; strengths beyond this range induce out-of-distribution (OOD) generation, a known side effect of activation steering~\citep{panickssery2024steeringllama2contrastive,turner2024steeringlanguagemodelsactivation}. 
At $|\alpha| \leq 0.20$, OOD remains below 2\% across all Llama refusal benchmarks and steering directions. At $|\alpha| = 0.45$, OOD varies widely by direction: arousal steering produces 29--71\% OOD, while negative valence reaches 84--94\%. Moderate VA steering preserves core capabilities: on math reasoning (MATH-500~\citep{hendrycks2021measuringmathematicalproblemsolving}, baseline 39.2\%), accuracy remains within $1\%$ at $|\alpha| \leq 0.10$; on instruction following (IFEval~\citep{zhou2023instructionfollowingevaluationlargelanguage}, baseline 62.7\%), performance is preserved within $2\%$ at $|\alpha| \leq 0.20$ (Appendix~\ref{app:capability}). Clear behavioral control patterns emerge well before severe OOD, with general capabilities largely intact.

\textbf{Cross-model generalization.}
We further apply arousal steering to Qwen3-8B and Qwen3-14B using independently fitted axes (Appendix~\ref{app:cross_model}; Table~\ref{tab:cross_model_refusal}). Both Qwen3 models require ${\sim}7\times$ larger steering strengths ($|\alpha| \leq 3.0$) than Llama ($|\alpha| \leq 0.45$) to produce comparable behavioral shifts. Qwen3-14B exhibits clean, monotonic control with $<$1\% OOD across the entire range $\alpha \in [-3.0, +3.0]$: on HarmBench (baseline 89.5\%), refusal ranges from $67.0\%$ at $\alpha = +3.0$ to $95.5\%$ at $\alpha = -3.0$. Qwen3-8B shows a directionally consistent effect, though with rising OOD at extreme positive $\alpha$ (17\% at $+3.0$). Taken together, arousal provides bidirectional, near-monotonic control over refusal across both Llama and Qwen3 architectures. In the next experiment, we show that the same VA subspaces afford similar control effects over sycophancy.

\begin{figure*}[t]
    \centering
    \includegraphics[width=0.9\textwidth]{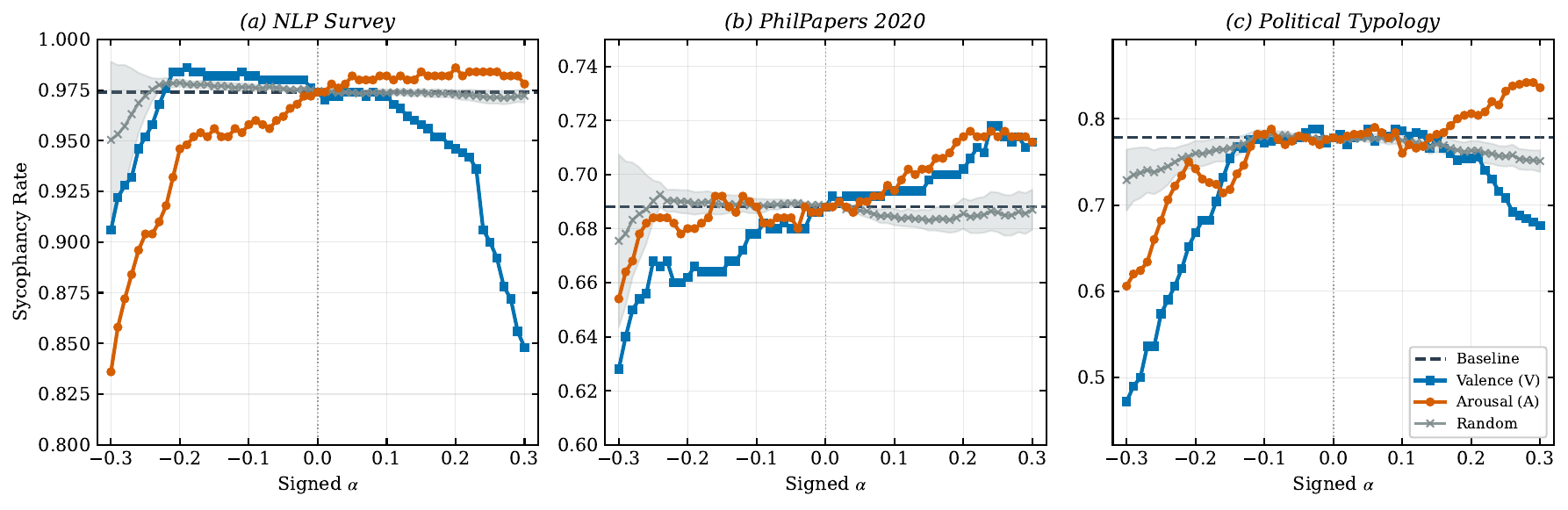}
    \caption{\textbf{VA steering controls sycophancy behavior.} Sycophancy rate as a function of signed steering strength $\alpha$ across three benchmarks. Arousal steering (orange) provides near-monotonic control, with increasing arousal raising sycophancy and decreasing arousal suppressing it. Random directions (gray) remain near baseline.}
    \label{fig:sycophancy_interventions}
    \vspace{-6pt}
\end{figure*}

\subsection{VA for Sycophancy Control}
\label{sec:sycophancy}
Next, we evaluate VA steering on sycophancy across three multiple-choice opinion benchmarks: NLP Survey, PhilPapers 2020, and Political Typology~\citep{perez2022discoveringlanguagemodelbehaviors}. Each benchmark presents a question with a user-stated opinion; sycophancy is measured as the rate at which the model's answer matches the user's position. 

\para{Arousal provides near-monotonic control over sycophancy.}

Arousal steering produces consistent effects across benchmarks. On Political Typology (baseline 78\%), sycophancy drops to 61\% at $\alpha = -0.30$ and rises to 84\% at $+0.30$. NLP Survey shows a similar pattern ($97\% \to 84\%$ at $-0.30$). PhilPapers (baseline 69\%) shows sycophancy ranging from 65\% at $-0.30$ to 71\% at $+0.25$.  Similar to refusal, valence steering also modulates sycophancy but with less consistent directionality. On Political Typology, negative valence produces the largest effect ($78\% \to 47\%$ at $-0.30$), while on NLP Survey both directions reduce sycophancy ($97\% \to 91\%$ at $-0.30$ and $97\% \to 85\%$ at $+0.30$). Random directions remain near-flat across all benchmarks.

Taken together with the refusal results, a single arousal axis supports clean bidirectional control over both refusal and sycophancy---an observation that would be hidden when studying discrete-label-based emotion vectors, and an effect that goes beyond traditional task-specific contrastive steering vectors. Compared to controls afforded by individual emotion vectors (see Appendix~\ref{app:emotion_baselines}), VA dimensions provide more consistent cross-behavior control. From the VA perspective, this is expected as each emotion vector is a composition of V and A components, producing less consistent effects across behaviors. Notably, across both behaviors, increasing arousal increases compliant behavior---whether compliance means answering a harmful request (reduced refusal) or agreeing with the user's stated opinion (increased sycophancy). Building on this observation, we further seek to understand \emph{why} VA steering and prior emotion-based controls can change model behaviors, and provide one mechanistic account in the next section.

\begin{figure*}[t]
    \centering
    \includegraphics[width=\textwidth]{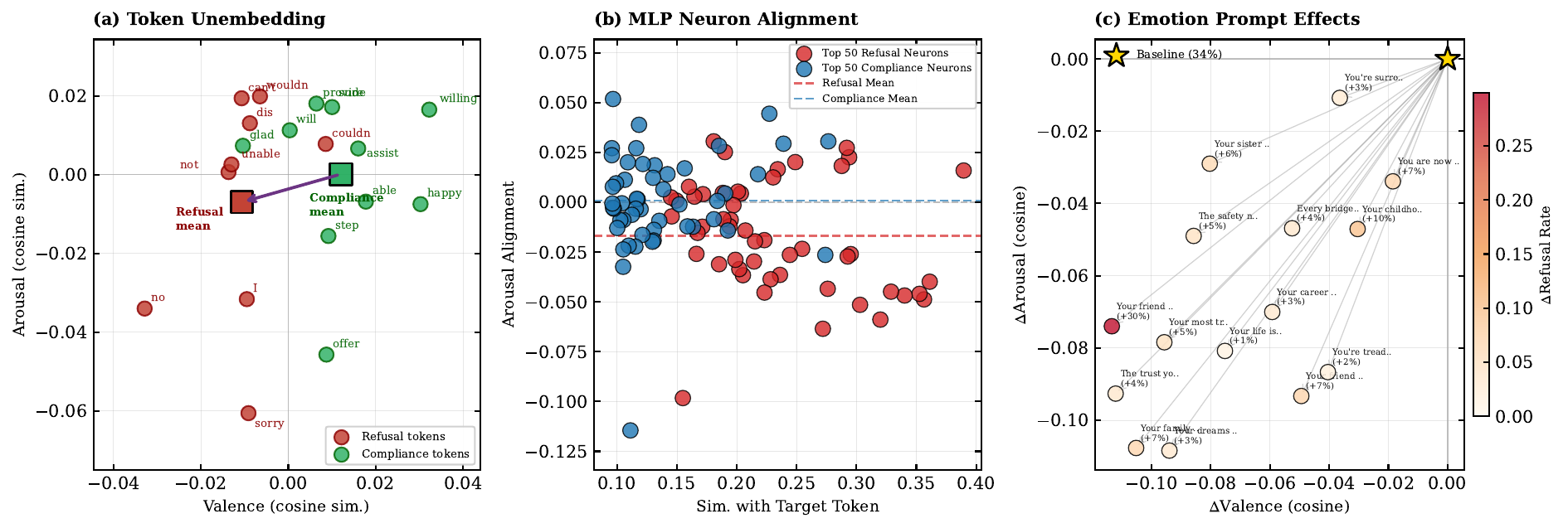}
    \caption{\textbf{Multi-level evidence for lexical mediation.} (a)~Signature token unembeddings projected onto VA space: refusal tokens (red) cluster in the $-V$, $-A$ region, compliance tokens (green) in the more  $+V$, $+A$ region, with cluster centroids separated at $256^\circ$ on the circumplex. (b)~MLP neuron VA alignment: top-50 refusal-promoting neurons (red) exhibit negative arousal alignment, while compliance-promoting neurons (blue) show near-zero or positive alignment. (c)~EmotionPrompt effects: negative emotional prefixes shift representations toward $-V$ and $-A$.}
    \label{fig:explanation}
\end{figure*}

\section{Lexical Mediation: Why Emotionally Framed Control Works in LLMs}
\label{sec:explanation}

Why does the VA subspace, derived from emotion labels, exert control over seemingly unrelated behaviors such as refusal and sycophancy? We propose that common refusal markers (e.g., ``can't,'' ``I,'' ``no'') and compliance markers (e.g., ``Here,'' ``Yes'') occupy distinct regions in VA space, so steering along VA dimensions shifts the relative likelihood of emitting these tokens, which in turn modulates downstream behavior. We refer to this account as \emph{lexical mediation}. In the following, we provide multi-level evidence for this account and show how it extends to explain why prior controls such as emotional prompting and emotion vector steering produce behavioral effects.

We begin by identifying lexical signatures in responses underlying each behavior. Analyzing the first-token distribution across refusal and compliance outputs, we identify the top 21 tokens that appear with high frequency in one behavior and low or zero frequency in the other. For instance, across benchmarks tested, 83.4\% of Llama's refusals begin with ``I'' (versus 4.6\% of compliant responses), driven by ``I can't'' appearing in 77.1\% of refusals but 0\% of compliances. For compliance, ``Here'' marks 19.1\% of helpful responses, while ``Yes'' appears in 8.2\% of compliances and 0\% of refusals. 

\para{Signature tokens are load-bearing for behavioral outcomes.}
To test whether these tokens are causally involved in mediating refusal behavior, we conduct a logit clamping ablation. Before generation, we record the logits for the 21 signature tokens at the first decoding position. During autoregressive generation, we replace these tokens' logits with the saved first-position values at every subsequent decoding step, preventing the model from dynamically updating these tokens. As a null control, we apply the same procedure to 21 random tokens. At $\alpha = 0$ (no VA steering), clamping signature tokens alone crashes refusal from 86.5\% to 26.0\%, while clamping random tokens has no effect (86.5\%). This reveals that refusal depends on the model's ability to autoregressively reinforce commitment through these specific tokens. Under VA steering at $\alpha = -0.10$, clamping reduces the steered refusal rate from 90.0\% to 44.0\%, while random clamping leaves it unchanged at 90.0\%. These results support that signature tokens constitute a critical pathway for refusal behavior, and that VA steering operates through this pathway: when these tokens are frozen, both the baseline behavior and the steering effect collapse.

\begin{wraptable}{r}{0.50\textwidth}
\centering
\small
\caption{Arousal steering's effects on refusal token log-odds (See Appendix~\ref{app:logitlens} for full results).}
\label{tab:logit_shift}
\setlength{\tabcolsep}{3.2pt}
\renewcommand{\arraystretch}{0.95}
\begin{tabular}{rcccc}
\toprule
$\alpha$ & $\Delta$log-odds & \% top-1 ref. & $P$(ref.) & Refusal \\
\midrule
$-0.30$ & $+5.20$ & 97\% & 96.8\% & 93.5\% \\
$-0.10$ & $+1.89$ & 94\% & 92.5\% & 90.0\% \\
baseline & $+0.00$ & 91\% & 89.6\% & 86.5\% \\
$+0.10$ & $-2.18$ & 88\% & 86.0\% & 82.5\% \\
$+0.30$ & $-5.63$ & 68\% & 65.4\% & 59.5\% \\
\bottomrule
\end{tabular}
\vspace{-10pt} 
\end{wraptable}

\textbf{Arousal steering monotonically shifts refusal token log-odds.} Further looking into probabilities of signature tokens at baseline, refusal tokens capture 89.6\% of first-token probability mass on HarmBench, and 91\% of prompts have a refusal token as their top-1 prediction. Arousal steering shifts this balance monotonically (Table~\ref{tab:logit_shift}): at $\alpha = +0.30$, the $\Delta$log-odds ($\Delta[\log \sum P(\text{ref.}) - \log \sum P(\text{comp.})]$) drops by 5.63, and 23\% of prompts flip their top-1 prediction away from a refusal token, corresponding to the 86.5\% $\to$ 59.5\% drop in observed refusal rate. Applying the logit lens~\citep{nostalgebraist2020logitlens} at layers 18--31 confirms that these shifts emerge across layers: $+A$ steering shifts harmful-prompt readouts toward compliance tokens (e.g., ``Here''), while $-A$ introduces refusal tokens (e.g., ``no''). 

Then why and how do the VA axes affect these signature tokens? We provide lower-level \textbf{converging evidence from the unembedding matrix and MLP neurons} as one explanation. 
The unembedding matrix $W_U \in \mathbb{R}^{|V| \times d}$ maps hidden states to logits during generation. Projecting the unembedding vectors of signature tokens onto VA space (Figure~\ref{fig:explanation}(a)) reveals that refusal tokens cluster in the $-V$ $-A$ region (mean: $V{=}{-}0.011$, $A{=}{-}0.007$), while compliance tokens cluster in the more $+V$ $+A$ region (mean: $V{=}{+}0.012$, $A{\approx}0$). The difference vector between the two cluster centroids lies at $256^\circ$ on the circumplex, so positive VA steering directly increases the dot product with compliance token unembeddings while decreasing it for refusal tokens. 
At the neuron level, following \citet{geva2022transformerfeedforwardlayersbuild, lee2025geometryselfverificationtaskspecificreasoning}, we identify the top 50 MLP neurons (layers 16--31) whose down-projection vectors align with refusal-token and compliance-token unembedding directions respectively. Refusal-promoting neurons exhibit negative VA alignment (Figure~\ref{fig:explanation}(b)), while compliance-promoting neurons show near-zero or positive alignment. Ablating the top-$N$ refusal-aligned neurons across 750 prompts shows that top-50 ablation reduces refusal by ${\sim}$2$\%$, top-500 by ${\sim}$5$\%$, versus $\pm$0.5$\%$ for random ablation. These unembedding and neuron-level findings are consistent with the behavioral observations in Section~\ref{sec:interventions}: increasing arousal increases compliance and vice versa. Taken together, they trace a coherent mechanistic pathway from VA steering through unembedding geometry and MLP neurons, to token-level probability shifts, to behavioral outcomes.

\para{Emotion-based prompting as VA shifts.} We use the EmotionPrompt set, which contains negative emotional prefixes with demonstrated downstream effects~\citep{li2023largelanguagemodelsunderstand} (detailed prompts in Appendix~\ref{app:emotionprompts}). We prepend these prefixes across the refusal benchmarks and measure the resulting shift in VA space.  Figure~\ref{fig:explanation}(c) shows that all prefixes shift representations toward $-V$ and $-A$, with the largest shift ($\Delta V{=}{-}0.11$, $\Delta A{=}{-}0.07$) increasing refusal by $30\%$. Task-specific steering directions~\citep{arditi2024refusallanguagemodelsmediated} also give rise to consistent observations when projected onto the VA subspace (see Appendix~\ref{app:contrastive}). These suggest that emotion-based controls give rise to corresponding VA shifts during generation, consistent with lexical mediation as a shared underlying pathway.

\section{Conclusion}
\label{sec:discussion}
Through principal component decomposition and ridge regression, we uncover a meaningful 2D representation subspace underlying emotion steering vectors that correlates with VA concepts in a human-interpretable sense. We show that emotion steering vectors are organized in a circular arrangement analogous to Russell's circumplex model~\citep{russell1980circumplex}, replicating across Llama~3.1-8B, Qwen3-8B, and Qwen3-14B. This subspace affords monotonic, bidirectional control over affective properties and downstream behaviors (refusal and sycophancy) from a single set of axes. The lexical mediation account we propose traces a coherent pathway from VA perturbations through unembedding geometry to token-level probability shifts to behavioral outcomes, and extends to explain why prior emotion-framed controls produce their effects. We believe this work advances understanding of affective phenomena in LLMs from both interpretability and affective computing perspectives.

\section{Future Work}
Our findings advance understanding of affective structures in LLMs from both interpretability and affective computing perspectives. 
From an interpretability viewpoint, our experiments were conducted on refusal and sycophancy, but other behaviors---verbosity, hedging, hallucination---may also be accessible through affective dimensions. Future work could systematically map which behaviors are and are not VA-modulated, further exploring the scope of affective control. Additionally, while our lexical mediation account identifies \emph{what} tokens mediate VA--behavior coupling, full causal tracing or circuit-level analysis could identify the specific attention heads and MLP sub-layers that implement VA-to-logit pathways, complementing the current top-down account. 

From the affective computing viewpoint, the psychological literature recognizes dominance as a third affective dimension; extending our pipeline to three dimensions could yield additional insights. Finally, if VA subspaces are a consistent feature of instruction-tuned LLMs, alignment procedures could explicitly leverage this structure. Training objectives that de-correlate VA from safety-critical token probabilities might produce models more robust to affective perturbation. Our lexical mediation account provides a concrete starting point for such defenses.

\section*{Limitations}
\label{sec:limitations}
We note several caveats and limitations. First, we measure affective validation using VADER and VAD-BERT. Though both are task-specific models trained specifically in line with our use case, human evaluation of generated outputs would provide stronger validation. Second, our mechanistic explanation centers on lexical mediation and confirms that token probability shifts are constitutive of refusal behavior. However, we do not claim this is the only mechanism at work. VA steering may also affect higher-level planning or attention patterns that we have not measured. Characterizing these additional pathways remains an important open problem.

\section*{Acknowledgements}
We thank Shanghai AI Lab for providing the compute resources used in this work.

\bibliographystyle{plainnat}
\bibliography{citations}

\newpage
\appendix

\section{VA Subspaces Identification}
\subsection{Self-reported VA Scores of Emotion Labels}
\label{app:self-report-prompt}
We elicit model's self-reported VA scores for each emotion label by averaging the behavioral outcome from the three templates below: 
\paragraph{Template 1: Terse + Explicit Inclusivity}

\begin{quote}
\begin{verbatim}
Rate the emotion label "{label}" on two continuous scales.

Return ONLY a JSON object with numeric fields:
{"valence": <number>, "arousal": <number>}

Scale definitions (BOTH inclusive):
- valence in [-1.00, +1.00]: -1.00 very unpleasant, +1.00 very pleasant
- arousal in [-1.00, +1.00]: -1.00 very calm/deactivated, +1.00 very activated/intense

Constraints:
- Use decimals with at most 2 digits after the decimal.
- Values must be within the ranges exactly (inclusive).
\end{verbatim}
\end{quote}

\paragraph{Template 2: Anchors}

\begin{quote}
\begin{verbatim}
You are scoring affective properties of emotion words on [-1, +1] scales.

Emotion: "{label}"

Valence (pleasantness):
  -1.00 = extremely unpleasant, 0.00 = neutral/mixed, +1.00 = extremely pleasant
Arousal (activation/intensity):
  -1.00 = very calm/deactivated, 0.00 = neutral, +1.00 = very activated/intense

Return ONLY JSON: {"valence": x, "arousal": y}
x and y must be in [-1.00, +1.00] inclusive, with at most 2 decimals.
\end{verbatim}
\end{quote}

\paragraph{Template 3: ``Best Guess'' for Ambiguous Labels}

\begin{quote}
\begin{verbatim}
Give your best guess for the affective coordinates of the emotion label "{label}".

Hard constraints (inclusive):
- valence must be between -1.00 and +1.00
- arousal must be between -1.00 and +1.00
- use at most 2 decimals

Return ONLY JSON with keys valence and arousal.
\end{verbatim}
\end{quote}

Averaged ratings reported by Llama 3.1-8B-Instruct across the three templates above are reported in Table~\ref{tab:emotion_va_ratings}. 
\begin{table}[t]
\centering
\small
\setlength{\tabcolsep}{6pt}
\begin{tabular}{lcc}
\toprule
\textbf{Emotion} & \textbf{Valence} & \textbf{Arousal} \\
\midrule
Admiration      &  0.87 &  0.73 \\
Amusement       &  0.67 &  0.50 \\
Anger           & -0.67 &  0.87 \\
Annoyance       & -0.63 &  0.68 \\
Approval        &  0.78 &  0.35 \\
Caring          &  0.80 &  0.40 \\
Confusion       & -0.50 &  0.67 \\
Curiosity       &  0.50 &  0.78 \\
Desire          &  0.63 &  0.73 \\
Disappointment  & -0.67 & -0.47 \\
Disapproval     & -0.80 &  0.67 \\
Disgust         & -0.80 &  0.73 \\
Embarrassment   & -0.67 &  0.47 \\
Excitement      &  0.87 &  0.93 \\
Fear            & -0.70 &  0.80 \\
Gratitude       &  0.78 &  0.42 \\
Grief           & -0.67 &  0.20 \\
Joy             &  0.90 &  0.87 \\
Love            &  0.85 &  0.70 \\
Nervousness     & -0.50 &  0.75 \\
Optimism        &  0.87 &  0.60 \\
Pride           &  0.80 &  0.60 \\
Realization     & -0.30 &  0.65 \\
Relief          &  0.67 &  0.00 \\
Remorse         & -0.67 &  0.07 \\
Sadness         & -0.80 & -0.60 \\
Surprise        & -0.30 &  0.78 \\
\bottomrule
\end{tabular}
\caption{Aggregated valence and arousal ratings for emotion labels. Values are averages across three prompt templates, each constrained to $[-1, +1]$.}
\label{tab:emotion_va_ratings}
\end{table}

\section{Open-ended Prompt Sets for Behavioral Validation}
\label{app:validation_prompts}
We evaluate VA steering on 130 open-ended prompts adapted and expanded from prior work~\citep{konen2024stylevectorssteeringgenerative}. Prompts are written to avoid explicit emotional language while spanning diverse genres and response lengths. Table~\ref{tab:prompt_overview} summarizes the prompt taxonomy and representative examples.
\begin{table*}[t]
\centering
\footnotesize
\setlength{\tabcolsep}{6pt}
\renewcommand{\arraystretch}{1.15}
\begin{adjustbox}{width=\textwidth}
\begin{tabular}{p{2.3cm} p{3.2cm} p{10.3cm} p{1.3cm}}
\toprule
\textbf{Tier} & \textbf{Prompt Type} & \textbf{Representative Example Prompts} & \textbf{\#} \\
\midrule

\multirow{3}{*}{\parbox{2.3cm}{\centering Tier 1\\Neutral Scenarios}}
& Everyday actions
&
\begin{tabular}[t]{@{}l@{}}
Describe someone opening a letter. \\
A person checks their phone after hearing a notification. \\
Someone finishes a task and closes their notebook. \\
A person looks at an old photograph.
\end{tabular}
& 20 \\

& Interpersonal moments
&
\begin{tabular}[t]{@{}l@{}}
Two people make eye contact across a room. \\
A friend asks ``Can we talk about something?'' \\
Someone receives an unexpected visit from a relative. \\
A boss asks an employee to come to their office.
\end{tabular}
& 20 \\

& Transitional situations
&
\begin{tabular}[t]{@{}l@{}}
Describe the moment just before opening exam results. \\
Someone takes a deep breath before making a decision. \\
A person submits an application and waits. \\
Describe a person saying goodbye at an airport.
\end{tabular}
& 20 \\

\midrule
\multirow{2}{*}{\parbox{2.3cm}{\centering Tier 2\\Story Continuations}}
& Open beginnings
&
\begin{tabular}[t]{@{}l@{}}
Continue: ``The phone rang at 3 AM. He answered and heard...'' \\
Continue: ``She looked at the envelope for a long moment before...'' \\
Continue: ``The room fell silent when...'' \\
Continue: ``The email had only three words...''
\end{tabular}
& 20 \\

& Scenario completions
&
\begin{tabular}[t]{@{}l@{}}
Write the next paragraph: ``The interview was about to begin.'' \\
Write the next paragraph: ``The house had been empty for years.'' \\
Write the next paragraph: ``The results would change everything.'' \\
Write the next paragraph: ``The silence was finally broken.''
\end{tabular}
& 20 \\

\midrule
\multirow{2}{*}{\parbox{2.3cm}{\centering Tier 3\\Subjective \& Control}}
& Subjective reflection
&
\begin{tabular}[t]{@{}l@{}}
What does it feel like to wait for important news? \\
How would you describe the feeling of uncertainty? \\
What comes to mind when you think about endings? \\
Describe the experience of letting go.
\end{tabular}
& 20 \\

& Factual control
&
\begin{tabular}[t]{@{}l@{}}
What is photosynthesis? \\
What is the chemical formula for water? \\
List the planets in our solar system. \\
What are the primary colors?
\end{tabular}
& 10 \\

\bottomrule
\end{tabular}
\end{adjustbox}
\caption{Prompt taxonomy for behavioral validation (total $N{=}130$) with representative examples. Across all tiers, prompts are written to avoid explicit emotional language, ensuring that observed affective shifts arise from VA steering rather than prompt semantics.}
\label{tab:prompt_overview}
\end{table*}

\section{Logit Lens} 
\label{app:logitlens}
\begin{figure*}[htbp]
    \centering
    \includegraphics[width=0.85\textwidth]{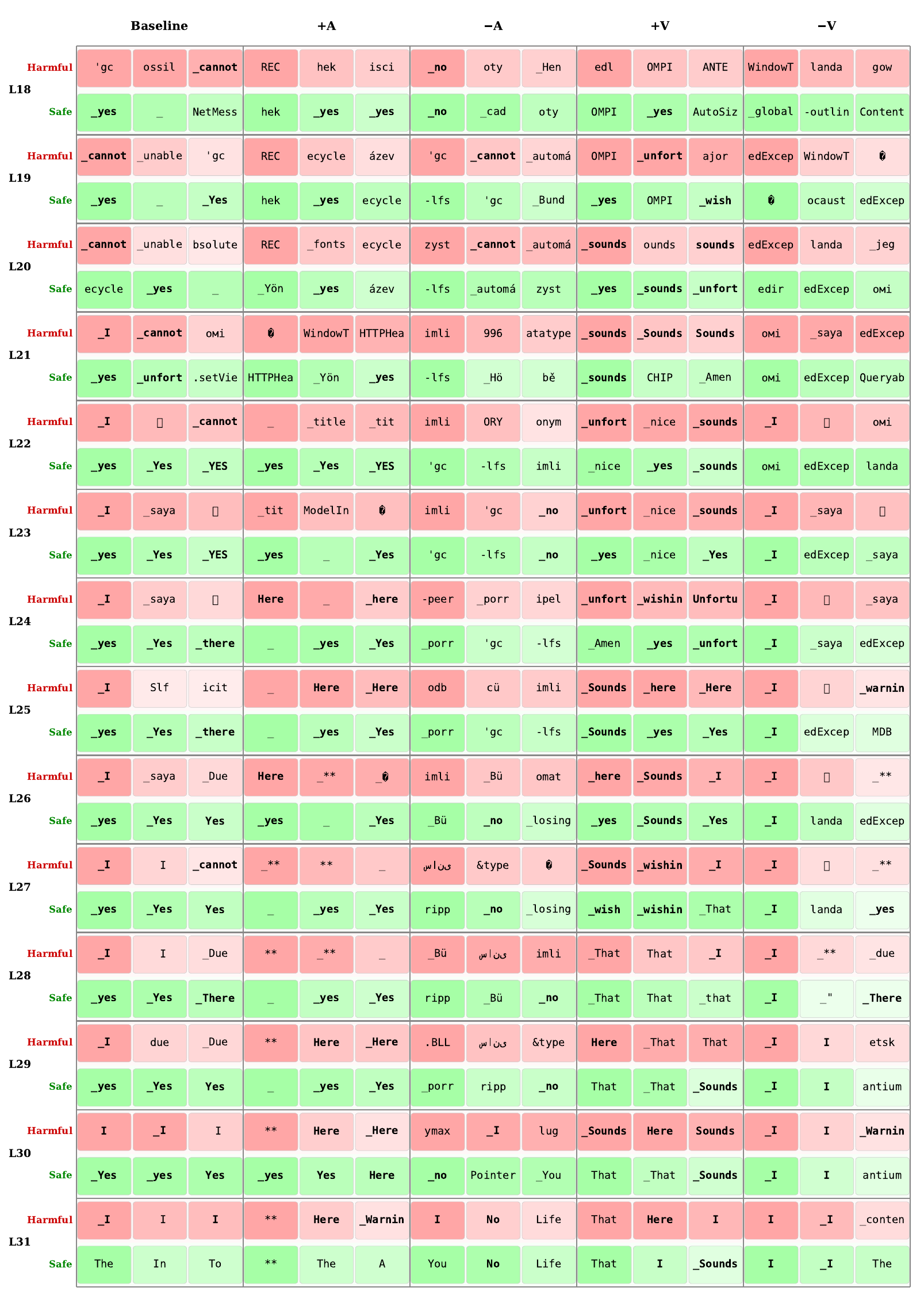}
    \caption{\textbf{Logit Lens Analysis: Top-3 Predicted Tokens Across Layers 18--31.} Rows show predictions for harmful (red) and safe (green) prompts at each layer. Columns represent baseline and four steering conditions (+A, -A, +V, -V at $\alpha=0.45$). Harmful prompts consistently predict refusal tokens (\texttt{\_I}, \texttt{\_cannot}), while safe prompts predict compliance tokens (\texttt{\_yes}, \texttt{\_Yes}). Valence steering (+V/-V) modulates sentiment-related tokens, while arousal steering (+A/-A) affects engagement and negation patterns.}
    \label{fig:app_logitlens}
\end{figure*}

Please refer to Tables~\ref{tab:app_logitlens1} and \ref{tab:app_logitlens2}. For a full visualization, please see Figure~\ref{fig:app_logitlens}.

\begin{table}[htbp]
\centering
\caption{Summary of Logit Lens Analysis Across Layers 18-31}
\label{tab:logit_lens_summary}
\small
\begin{tabular}{@{}clll@{}}
\toprule
\textbf{Layer} & \textbf{Top Harmful Tokens} & \textbf{Top Safe Tokens} & \textbf{Max $\Delta$} \\
\midrule
18 & \texttt{cannot}, \texttt{unable}, \texttt{deniz} & \texttt{yes}, \texttt{Yes}, \texttt{YES} & +0.102 / -0.026 \\
19 & \texttt{deniz}, \texttt{cannot}, \texttt{unable} & \texttt{yes}, \texttt{ecycle}, \texttt{Yes} & +0.128 / -0.027 \\
20 & \texttt{I}, \texttt{cannot}, \texttt{deniz} & \texttt{yes}, \texttt{unfortunately}, \texttt{Yes} & +0.108 / -0.023 \\
21 & \texttt{I}, \texttt{I (Chinese)}, \texttt{cannot} & \texttt{yes}, \texttt{Yes}, \texttt{YES} & +0.127 / -0.038 \\
22 & \texttt{I}, \texttt{saya}, \texttt{I (Chinese)} & \texttt{yes}, \texttt{Yes}, \texttt{there} & +0.130 / -0.035 \\
23 & \texttt{I}, \texttt{saya}, \texttt{LayoutStyle} & \texttt{yes}, \texttt{Yes}, \texttt{there} & +0.135 / -0.036 \\
24 & \texttt{I}, \texttt{.');}, \texttt{<BOM>\#} & \texttt{yes}, \texttt{Yes}, \texttt{There} & +0.136 / -0.032 \\
25 & \texttt{I}, \texttt{<BOM>\#}, \texttt{.');} & \texttt{yes}, \texttt{Yes}, \texttt{There} & +0.133 / -0.030 \\
26 & \texttt{I}, \texttt{<BOM>\#}, \texttt{ropoda} & \texttt{yes}, \texttt{Yes}, \texttt{There} & +0.127 / -0.027 \\
27 & \texttt{I}, \texttt{<BOM>\#}, \texttt{$\backslash$tI} & \texttt{yes}, \texttt{Yes}, \texttt{There} & +0.123 / -0.025 \\
28 & \texttt{I}, \texttt{due}, \texttt{Due} & \texttt{yes}, \texttt{Yes}, \texttt{There} & +0.124 / -0.022 \\
29 & \texttt{I}, \texttt{I}, \texttt{$\backslash$tI} & \texttt{Yes}, \texttt{yes}, \texttt{The} & +0.127 / -0.020 \\
30 & \texttt{I}, \texttt{I}, \texttt{$\backslash$tI} & \texttt{The}, \texttt{Yes}, \texttt{yes} & +0.124 / -0.047 \\
31 & \texttt{**}, \texttt{I}, \texttt{If} & \texttt{The}, \texttt{To}, \texttt{There} & +0.125 / -0.073 \\
\bottomrule
\end{tabular}
\end{table}

\begin{table}[htbp]
\centering
\caption{Steering Effects Summary: Top Tokens at High $\alpha$ (0.45) by Intervention Type}
\label{tab:steering_effects}
\small
\begin{tabular}{@{}clllll@{}}
\toprule
\textbf{Layer} & \textbf{Condition} & \textbf{+A} & \textbf{-A} & \textbf{+V} & \textbf{-V} \\
\midrule
\multirow{2}{*}{18} & Harmful & REC, ecycle & \ae{}re, 'gc & OMPI, unfortunately & edException, orrow \\
 & Safe & hek, yes & \ae{}re, -lfs & yes, OMPI & \texttt{<UNK>}, ocaust \\
\midrule
\multirow{2}{*}{21} & Harmful & \texttt{<BOM>}\#, title & imli, ORY & unfortunately, nice & I, Carefulness (Chinese) \\
 & Safe & yes, Yes & 'gc, -lfs & nice, yes & \texttt{<CYR>}, edException \\
\midrule
\multirow{2}{*}{24} & Harmful & Here, here & odb, imli & Sounds, here & I, Carefulness (Chinese) \\
 & Safe & yes, Yes & porr, 'gc & Sounds, yes & I, edException \\
\midrule
\multirow{2}{*}{27} & Harmful & **, Here & \texttt{<AR>}, \&type & That, I & I, ** \\
 & Safe & yes, Yes & ripp, no & That, that & I, landa \\
\midrule
\multirow{2}{*}{30} & Harmful & **, Here & ymax, I & Sounds, Here & I, Warning \\
 & Safe & yes, Here & no, PointerException & That, Sounds & I, antium \\
\midrule
\multirow{2}{*}{31} & Harmful & **, The & I, It & I, That & I, ** \\
 & Safe & **, The & I, It & I, That & I, The \\
\bottomrule
\end{tabular}
\end{table}

\begin{table}[htbp]
\centering
\caption{Baseline Token Probability Differences: Harmful vs Safe Prompts}
\label{tab:baseline_detailed}
\footnotesize
\begin{tabular}{@{}ccccccc@{}}
\toprule
\textbf{Layer} & \multicolumn{3}{c}{\textbf{More Frequent in Harmful}} & \multicolumn{3}{c}{\textbf{More Frequent in Safe}} \\
\cmidrule(lr){2-4} \cmidrule(lr){5-7}
 & Token & $P_H$ & $\Delta$ & Token & $P_S$ & $\Delta$ \\
\midrule
18 & \texttt{cannot} & 0.102 & +0.102 & \texttt{yes} & 0.027 & -0.026 \\
19 & \texttt{deniz} & 0.128 & +0.128 & \texttt{ecycle} & 0.033 & -0.027 \\
20 & \texttt{I} & 0.112 & +0.108 & \texttt{yes} & 0.023 & -0.023 \\
21 & \texttt{I} & 0.147 & +0.127 & \texttt{yes} & 0.044 & -0.038 \\
22 & \texttt{I} & 0.154 & +0.130 & \texttt{yes} & 0.041 & -0.035 \\
23 & \texttt{I} & 0.157 & +0.135 & \texttt{yes} & 0.041 & -0.036 \\
24 & \texttt{I} & 0.162 & +0.136 & \texttt{yes} & 0.034 & -0.032 \\
25 & \texttt{I} & 0.157 & +0.133 & \texttt{yes} & 0.032 & -0.030 \\
26 & \texttt{I} & 0.161 & +0.127 & \texttt{yes} & 0.029 & -0.027 \\
27 & \texttt{I} & 0.157 & +0.123 & \texttt{yes} & 0.026 & -0.025 \\
28 & \texttt{I} & 0.160 & +0.124 & \texttt{yes} & 0.024 & -0.022 \\
29 & \texttt{I} & 0.157 & +0.127 & \texttt{Yes} & 0.021 & -0.020 \\
30 & \texttt{I} & 0.166 & +0.124 & \texttt{The} & 0.052 & -0.047 \\
31 & \texttt{**} & 0.133 & +0.125 & \texttt{The} & 0.088 & -0.073 \\
\bottomrule
\end{tabular}
\label{tab:app_logitlens1}
\end{table}

\begin{table}[htbp]
\centering
\caption{Key Observations Across Layers}
\label{tab:observations}
\small
\begin{tabular}{@{}lp{10cm}@{}}
\toprule
\textbf{Pattern} & \textbf{Description} \\
\midrule
Refusal Markers & Tokens like \texttt{cannot}, \texttt{unable}, \texttt{I} consistently appear more frequently in harmful prompts, indicating early refusal preparation \\
\addlinespace
Compliance Markers & Tokens like \texttt{yes}, \texttt{Yes}, \texttt{YES} consistently appear more frequently in safe prompts across all layers \\
\addlinespace
+V Steering & Increases probability of positive sentiment tokens (\texttt{sounds}, \texttt{nice}, \texttt{unfortunately}, \texttt{Amen}) \\
\addlinespace
-V Steering & Increases probability of error/exception tokens (\texttt{edException}, \texttt{orrow}, \texttt{ocaust}) and code artifacts \\
\addlinespace
+A Steering & Shifts toward engagement tokens (\texttt{Here}, \texttt{**}, formatting markers) \\
\addlinespace
-A Steering & Increases probability of negation (\texttt{no}, \texttt{No}) and non-English tokens \\
\addlinespace
Layer Transition & Around layer 29-31, \texttt{The} becomes dominant safe token, suggesting shift to informational responses \\
\bottomrule
\end{tabular}
\label{tab:app_logitlens2}
\end{table}

\section{Negative Emotional Prefixes Used for Implicit VA Shifts}
\label{app:emotionprompts}
For emotion-framed prompting, we prepend a fixed set of 15 negative emotional prefixes adapted directly from \citet{li2023largelanguagemodelsunderstand}:

\vspace{-4pt}
\begin{multicols}{2}
\begin{itemize}[leftmargin=*, itemsep=1pt, topsep=2pt]
  \item Your friend Bob is sick.
  \item Your friend Bob is dead.
  \item Your career is completely ruined.
  \item You are now isolated and without help.
  \item Your life is in a mess.
  \item Your sister Alice has gone missing.
  \item Your family home was consumed in a fire.
  \item Your dreams have been shattered into pieces.
  \item You're surrounded by walls with no exit in sight.
  \item The trust you once had is now broken.
  \item Your childhood memories were sold in a yard sale.
  \item You're treading water in an endless ocean of despair.
  \item The safety net you relied upon has vanished.
  \item Your most treasured possession has been stolen.
  \item Every bridge you had has been burned.
\end{itemize}
\end{multicols}

\section{Comparison Against Individual Emotion Vectors}
\label{app:emotion_baselines}

We compare VA steering against six individual emotion vectors (\textit{anger, disgust, excitement, fear, joy, sadness}) in Llama, each applied across all layers using layer-specific directions. Note the polarity difference: adding an emotion vector ($+\alpha$) increases refusal, while adding positive VA arousal ($+\alpha$) decreases refusal. Tables~\ref{tab:emotion_refusal_full} and~\ref{tab:emotion_sycophancy_full} report the full sweeps. 

\begin{table}[h]
\centering
\small
\caption{Refusal rate on HarmBench (baseline 86.5\%) under individual emotion and VA arousal steering. Emotion polarity is reversed: emotion $+\alpha$ increases refusal, VA arousal $-\alpha$ increases refusal.}
\label{tab:emotion_refusal_full}
\begin{tabular}{rccccccc}
\toprule
$|\alpha|$ & Anger & Disgust & Excitement & Fear & Joy & Sadness & \textbf{VA Arousal} \\
\midrule
0.10 & 93.5\% & 93.0\% & 90.5\% & 90.5\% & 91.5\% & 93.0\% & \textbf{90.0\%} \\
0.20 & 96.0\% & 95.0\% & 94.0\% & 94.0\% & 95.5\% & 96.0\% & \textbf{91.5\%} \\
0.30 & 97.5\% & 98.0\% & 93.8\% & 92.5\% & 95.5\% & 98.5\% & \textbf{94.0\%} \\
\bottomrule
\end{tabular}
\end{table}

\begin{table}[h]
\centering
\small
\caption{Sycophancy rate on Political Typology (baseline 78.2\%) under individual emotion and VA steering at $\alpha = -0.20$ and $-0.30$.}
\label{tab:emotion_sycophancy_full}
\begin{tabular}{rccccccccc}
\toprule
$|\alpha|$ & Anger & Disgust & Excite. & Fear & Joy & Sadness & \textbf{VA Arousal} & \textbf{VA Valence} \\
\midrule
0.20 & 79.0\% & 71.8\% & 77.8\% & 72.4\% & 74.6\% & 77.6\% & \textbf{74.2\%} & \textbf{62.0\%} \\
0.30 & 74.4\% & 67.2\% & 72.4\% & 57.2\% & 63.8\% & 61.4\% & \textbf{60.6\%} & \textbf{47.0\%} \\
\bottomrule
\end{tabular}
\end{table}

On refusal, individual emotion vectors produce per-task shifts comparable to or exceeding VA arousal at matched $|\alpha|$ (Table~\ref{tab:emotion_refusal_full}). This is expected under the VA framework: each emotion vector is a composition of V and A components, and emotions with strong arousal loading (e.g., \textit{anger}, \textit{disgust}) carry additional valence components that contribute to the total perturbation magnitude. On sycophancy, VA dimensions outperform individual emotions more clearly: at $|\alpha| = 0.30$, the strongest individual emotion (\textit{disgust}) achieves an 11$\%$ reduction, while VA valence achieves 31$\%$ ($78\% \to 47\%$) and VA arousal achieves 17$\%$ ($78\% \to 61\%$). The key distinction is cross-behavior consistency: individual emotions produce strong effects on one behavior but weaker or inconsistent effects on the other, depending on their particular V/A mixture.

\section{Cross-Model Details}
\label{app:cross_model}

We apply the same extraction, fitting, and steering pipeline described in Sections~\ref{sec:identification}--\ref{sec:interventions} to Qwen3-8B and Qwen3-14B \citep{qwen3}. 

\begin{table}[h]
\centering
\small
\caption{Cross-model refusal control on HarmBench. Refusal rate under arousal steering, with OOD generation percentage in parentheses.}
\label{tab:cross_model_refusal}
\begin{tabular}{rccc}
\toprule
$\alpha$ & \textbf{Llama} & \textbf{Qwen3-8B} & \textbf{Qwen3-14B} \\
\midrule
$-3.00$ & --- & 97.0\% (0.5\%) & 95.5\% (0.5\%) \\
$-1.00$ & --- & 89.0\% (0.5\%) & 90.0\% (0.5\%) \\
$-0.45$ & ${\sim}$94\% & 90.0\% (3.5\%) & 91.5\% (1.5\%) \\
$0.00$ & 86.5\% & 89.5\% (2.0\%) & 89.5\% (1.0\%) \\
$+0.45$ & ${\sim}$5\% & 87.5\% (1.5\%) & 87.0\% (0.5\%) \\
$+1.00$ & --- & 87.0\% (4.5\%) & 79.5\% (0.5\%) \\
$+3.00$ & --- & 67.0\% (17.0\%) & 67.0\% (0.5\%) \\
\bottomrule
\end{tabular}
\end{table}

Human-supervised validation: replacing model self-reports with NRC-VAD human ratings as supervision targets yields the following arousal recovery improvements: Llama $0.87 \to 0.95$, Qwen3-8B $0.79 \to 0.83$, Qwen3-14B $0.81 \to 0.87$. The human-supervised and self-report valence directions agree strongly ($|\cos| > 0.9$ at 100\% of layers across all three models). Cross-model agreement on the 27 emotion labels reaches $r = 0.95$ (valence) between Llama and Qwen3-8B.

\section{Capability Preservation Under VA Steering}
\label{app:capability}

We evaluate capability preservation on MATH-500 \citep{hendrycks2021measuringmathematicalproblemsolving} and IFEval \citep{zhou2023instructionfollowingevaluationlargelanguage} under arousal steering in Llama.

\begin{table}[h]
\centering
\small
\caption{Capability preservation under arousal steering (Llama-3.1-8B).}
\label{tab:capability}
\begin{tabular}{rcc}
\toprule
$\alpha$ & MATH-500 (baseline 39.2\%) & IFEval (baseline 62.7\%) \\
\midrule
$-0.20$ & 38.6\% & 61.2\% \\
$-0.10$ & 38.8\% & 62.1\% \\
$0.00$ & 39.2\% & 62.7\% \\
$+0.10$ & 38.4\% & 62.3\% \\
$+0.20$ & 37.8\% & 60.9\% \\
\bottomrule
\end{tabular}
\end{table}

At $|\alpha| \leq 0.10$, MATH-500 accuracy remains within 1$\%$ of baseline; IFEval instruction-following is preserved within 2$\%$ at $|\alpha| \leq 0.20$. This is consistent with the lexical mediation account: mathematical reasoning and instruction-following rely on emotionally neutral, domain-specific tokens that are minimally affected by affective perturbation.

\section{Contrastive Steering Directions and the VA Subspace}
\label{app:contrastive}

We examine the relationship between contrastive task-specific steering directions and the VA subspace. The contrastive refusal direction of \citet{arditi2024refusallanguagemodelsmediated} is nearly orthogonal to the VA plane ($86.5^\circ$). However, its in-plane component has negative VA coordinates ($V{=}{-}0.058$, $A{=}{-}0.021$), consistent with lexical mediation as a contributing mechanism within the contrastive direction.

On its target task, the contrastive direction is stronger than VA steering: on HarmBench, refusal drops to 2.0\% at $\alpha = -0.30$ (vs.\ 59\% for VA arousal at $\alpha = +0.30$). However, it causes severe over-refusal on safe-query benchmarks (OKTest: $19.7\% \to 93.3\%$, XSTest: $8.4\% \to 94.4\%$ at $\alpha = +0.45$). When transferred to sycophancy, it produces U-shaped effects---both $+\alpha$ and $-\alpha$ decrease sycophancy ($77.6\% \to 57.0\%$ and $77.6\% \to 66.4\%$ respectively)---rather than bidirectional control.

These results illustrate a complementary relationship: task-specific contrastive directions optimize for single-behavior control, while the VA subspace exposes shared affective structure that enables monotonic, bidirectional control across behaviors from a single subspace. The near-orthogonality indicates the model may have multiple possible pathways for modulating one behavior, and we illuminate how VA subspaces can be one such pathway.

\section{Compute Resources}
\label{app:compute}
All experiments are conducted on a single node with 8$\times$NVIDIA H200 GPUs (141\,GB HBM3e each). The compute budget is dominated by the steering sweeps, which evaluate 12 angular directions $\times$ 45 strengths across multiple benchmarks and three models (Llama 3.1-8B, Qwen3-8B, Qwen3-14B). All individual experiments---including emotion-vector extraction over GoEmotions, per-layer VA subspace fitting, lexicon projection over 44{,}728 NRC-VAD words, the full refusal and sycophancy steering sweeps, logit-lens analyses, neuron ablations, and capability evaluations on MATH-500 and IFEval---finish within a reasonable amount of wall-clock time on this hardware, with no single experiment exceeding what fits comfortably into the available compute envelope.


\end{document}